\documentclass{article}

\usepackage{microtype}
\usepackage{graphicx}
\usepackage{subfigure}
\usepackage{booktabs} 
\usepackage{algpseudocode}  

\usepackage[accepted]{icml2021_arxiv}

\icmltitlerunning{}

\algnewcommand\algorithmicforeach{\textbf{foreach}}
\algdef{S}[FOR]{ForEach}[1]{\algorithmicforeach\ #1\ \algorithmicdo}

\begin{document}

\twocolumn[
\icmltitle{Learning from Distinctive Candidates to Optimize \\
Reduced-Precision Convolution Program on Tensor Cores}

\begin{icmlauthorlist}
\icmlauthor{Junkyeong Choi}{nota}
\icmlauthor{Hyucksung Kwon}{hyu}
\icmlauthor{Woongkyu Lee}{hyu}
\icmlauthor{Jungwook Choi}{hyu}
\icmlauthor{Jieun Lim}{nota}

\end{icmlauthorlist}

\icmlaffiliation{nota}{NOTA, Korea}
\icmlaffiliation{hyu}{Artificial Intelligence Hardware \& Algorithms (AIHA) Lab., Hanyang University, Korea}

\icmlcorrespondingauthor{Junkyeong Choi}{junkyeong.choi@nota.ai}
\icmlcorrespondingauthor{Jieun Lim}{jieun.lim@nota.ai}

\vskip 0.3in
]

\printAffiliationsAndNotice{}  

\begin{abstract}
Convolution is one of the fundamental operations of deep neural networks with demanding matrix computation. In a graphic processing unit (GPU), Tensor Core is a specialized matrix processing hardware equipped with reduced-precision matrix-multiply-accumulate (MMA) instructions to increase throughput. However, it is challenging to achieve optimal performance since the best scheduling of MMA instructions varies for different convolution sizes. In particular, reduced-precision MMA requires many elements grouped as a matrix operand, seriously limiting data reuse and imposing packing and layout overhead on the schedule. This work proposes an automatic scheduling method of reduced-precision MMA for convolution operation. In this method, we devise a search space that explores the thread tile and warp sizes to increase the data reuse despite a large matrix operand of reduced-precision MMA. The search space also includes options of register-level packing and layout optimization to lesson overhead of handling reduced-precision data. Finally, we propose a search algorithm to find the best schedule by learning from the distinctive candidates. This reduced-precision MMA optimization method is evaluated on convolution operations of popular neural networks to demonstrate substantial speedup on Tensor Core compared to the state of the arts with shortened search time.
\end{abstract}


\section{Introduction}

Convolution is one of the fundamental operations of deep neural networks. Starting from the original work of \cite{alexnet2021}, convolution has been used as one of the core operations of convolutional neural networks\cite{krizhevsky2012imagenet,he2016deep,szegedy2016rethinking}. More recently, convolution was incorporated in the other types of neural networks such as recurrent neural network (RNN) and Transformer~\cite{dosovitskiy2020image} to collaborate within various network structures for capturing useful features in the computer vision domain. 

Although its versatility in various neural networks, convolution demands a large amount of computation and memory space. It is often computed Toeplitz matrix multiplication with dimension often spanning to hundreds to even thousands, taking up a significant portion of execution time. To alleviate the computational burden of convolution, various customized hardwares have been introduced. In a graphic processing unit (GPU), Tensor Core is a specialized matrix processing hardware equipped with reduced-precision matrix-multiply-accumulate (MMA) instructions to increase throughput. MMA instructions enable massively parallel execution of matrix multiplications grouped with a size of matrix operand increased inversely proportional to the bit-precision. For example, NVIDIA T4's INT4 MMA takes a group of 8x32 elements as an operand, which is twice larger than INT8 MMA (8x16)~\cite{carrasco2018analyzing}. These reduced MMA instructions are designed to provide a significant increase of throughput in executing reduced-precision convolution.

However, it is challenging to use reduced precision MMA instructions of Tensor Cores to achieve optimal performance in convolution execution. With a careful analysis, we identified three main sources of inefficiency in reduced precision MMA when it is used for representative convolution computations. First, the data reuse opportunity is significantly limited since MMA requires many elements grouped as matrix operands. Such a large grouping often causes unbalancing in workload division when setting up blocks and tiles for GPU execution. Second, reduced precision input should be packed before it is executed as operands of reduced precision MMA. In the case of INT4 MMA, the packing includes quantization of 8 consecutive values (in 32-bit) into a packed vector of 4-bit elements. This low-level data alignment incurs noticeable overhead with additional memory accesses. Third, due to this packing, the data layout of the convolution operand should be adjusted in a way that its lowest dimension matches with the packing granularity. Since the output of one convolution layer is often used as the input of the next convolution layer, mismatch of data layout between the output and input results in high re-layout cost due to uncoalesced memory accesses. 

One might overcome these causes of inefficiency by finding optimal scheduling of MMA instructions such as setting up of tile and block sizes that maximizes the data reuse while suppressing the overhead of data packing and re-layout. However, the best scheduling of MMA instructions varies for different convolution sizes. Not alone the variety of neural network structures, various shapes and sizes of convolution operations are incorporated even within a single neural network. For example, ResNet18 includes four stages of convolution layers, each of which takes different feature map sizes and the number of channels. Since the shape of input and weight imposes significantly different computation behavior for a convolution operation, it is not feasible to generalize a certain MMA scheduling for achieving universally optimal speedup.    

This work proposes an automatic scheduling method of reduced-precision MMA for convolution operation. In this method, we devise a search space that explores the thread tile and warp sizes to increase the data reuse despite a large matrix operand of reduced-precision MMA. The search space also includes options of register-level packing and layout optimization to lesson overhead of handling reduced-precision data. Finally, we propose a search algorithm to find the best schedule by learning from the distinctive candidates. This reduced-precision MMA optimization method is evaluated on convolution operations of popular neural networks to demonstrate substantial speedup on Tensor Core compared to the state of the arts with shortened search time.


\section{Background}

\subsection{Reduced Precision Convolution on Tensor Cores}

\begin{figure}[t]
\vskip 0in
\begin{center}
\centerline{\includegraphics[width=\columnwidth]{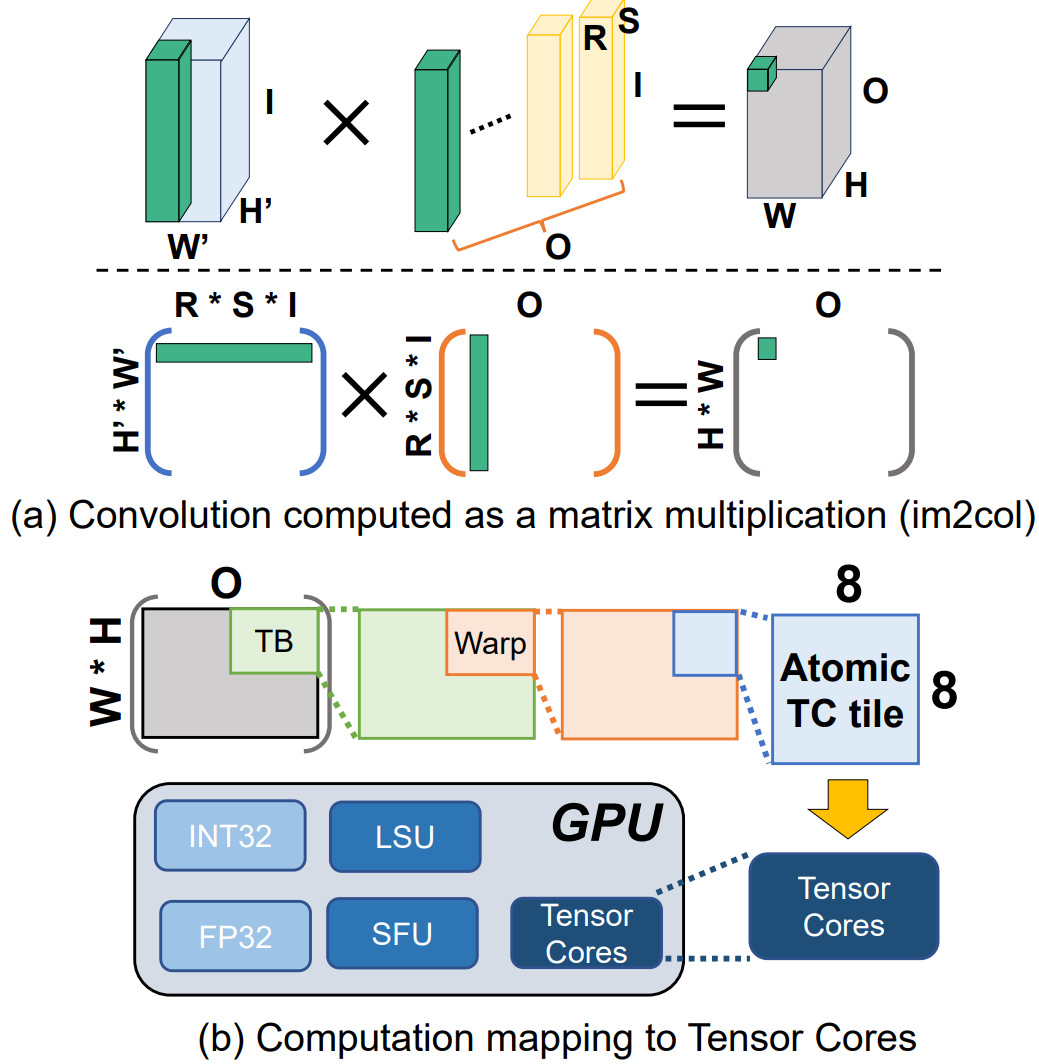}}
\caption{Reduced-Precision Convolution on Tensor Cores. Atomic WMMA tile size is 8x8.}
\label{conv2gemm}
\end{center}
\vskip -0.4in
\end{figure}

As a convolution consists of scalar multiplications and their accumulation, it is mathematically identical with vector inner product (Figure~\ref{conv2gemm}. Therefore, we can convert convolution into GEMM(General Matrix to Matrix Multiplication) with some data layout transformation called \emph{im2col}. Convolution with feature map width \textbf{\emph{W}}, height \textbf{\emph{H}}, input channel \textbf{\emph{I}}, output channel \textbf{\emph{O}}, and convolution kernel \textbf{\emph{R}}, \textbf{\emph{S}} with batch size of \textbf{\emph{N}} can be translated into matrix multiplication of \textbf{\emph{(N*H*W, I*R*S) $\times$ (I*R*S, O)}} where \textbf{\emph{N*H*W}} is the number of rows of the output matrix, \textbf{\emph{O}} is the number of columns and \textbf{\emph{I*R*S}} is the accumulation dimension.

In order to redesign convolution into a simple matrix multiplication problem, the input feature map is lowered into \textbf{im2col layout}. As Figure~\ref{conv2gemm}-(a) presents, every data required to process 3x3 convolution for each pixel are converted into a single matrix row. This process is also known as \textbf{lowering} as multi-dimensional tensors are converted into lower-level, two-dimensional matrix for efficient hardware implementation.

Enormous matrix multiplication can be parallelized using the matrix tiling procedure. As every element inside each row and column can be computed independently, each computation of the output matrix sub-title can be fully parallelized without any dependency among themselves. GPU, hardware specialized for parallelized workload, can exploit this nature of matrix multiplication. Furthermore, as all elements inside the output matrix can be parallelized, GPU can select any parallelization configuration. Figure~\ref{conv2gemm}-(b) depicts one example of parallel processing of tiled matrix multiplication. The output matrix is divided into multiple thread block tiles and then into thread warp tiles. Finally, each thread warp tile is divided into the smallest atomic WMMA(warp matrix-multiply-accumulate) tile with predefined size for Tensor Core instructions.

\subsection{Searching for Computation Schedule }

\begin{figure}[t]
\vskip 0.0in
\begin{center}
\centerline{\includegraphics[width=\columnwidth]{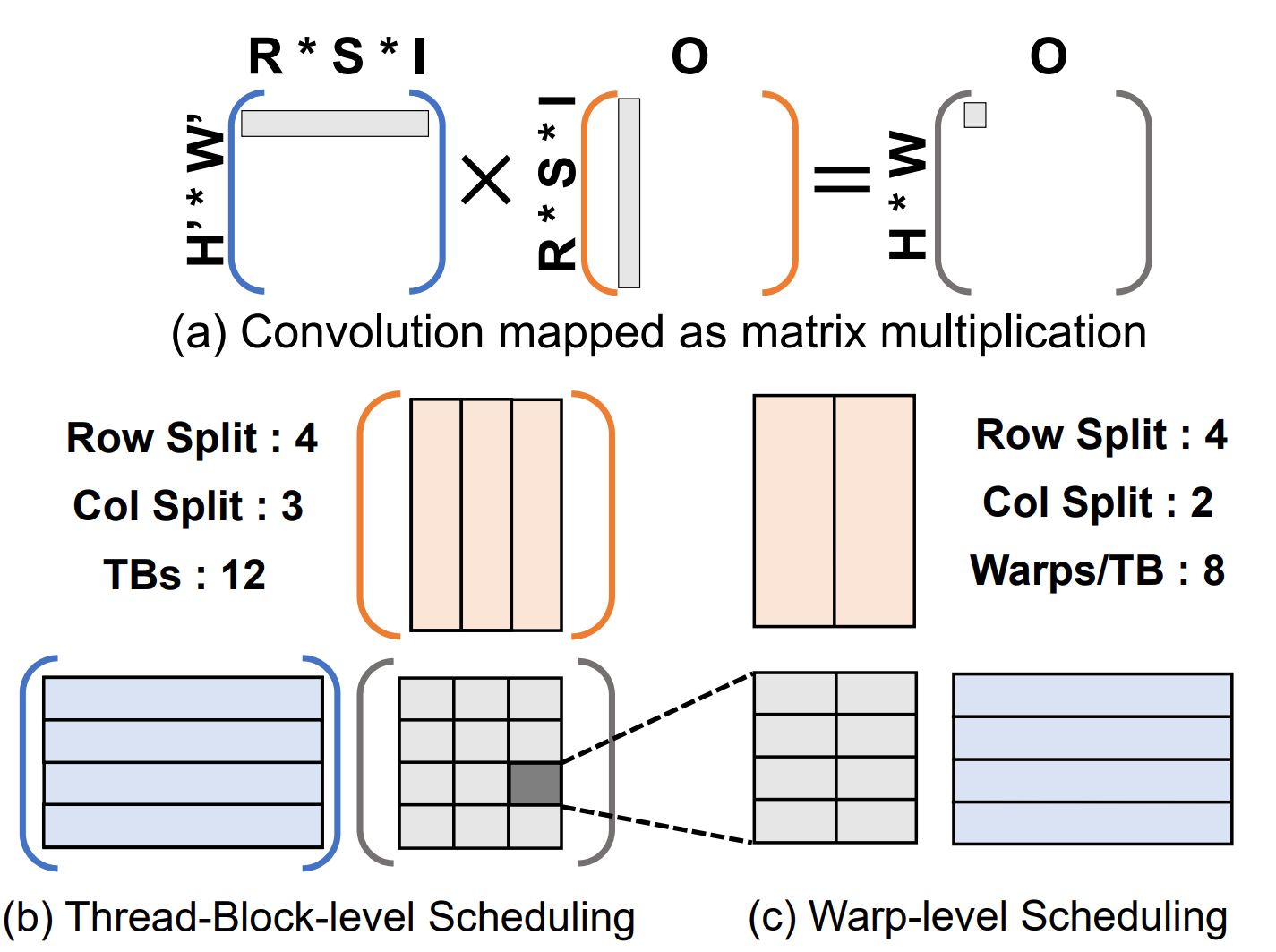}}
\caption{One example result of a search for optimal scheduling of MMA on Tensor Cores.}
\label{gemm_schedule}
\end{center}
\vskip -0.4in
\end{figure}

As GPUs can have a massive space of parallelization options, finding the optimal tile size for parallelism is critical to the overall runtime of the neural networks. Figure~\ref{gemm_schedule}-(b),(c) depicts one example of such parallelization scheduling. Too small parallelization may result in occupancy problems or inefficiency due to resource starvation. At the same time, too much parallelization may result in the serialization of workload or inefficiency due to the parallelization limit. Furthermore, we need to balance row-wise and column-wise parallelization because excessive parallelization on one side may result in excessive data load on the other side. In this sense, optimizing parallelization of convolution is challenging because the optimal parallelization option would depend on 1)~high-level convolution definition such as feature map width, height, input channels, output channels, convolution kernel size, batch size as well as 2)~GPU architecture and specification such as compute capability, the number of streaming multiprocessors(SMs), L1/L2 cache size, or processor performance.
We search this massive search space using modified AutoTVM with our contribution to the search algorithm.

\section{Method}

\subsection{Duplicate-Aware Load}

\subsubsection{Im2col Lowering}

\begin{figure}[t]
\vskip 0.1in
\begin{center}
\centerline{\includegraphics[width=\columnwidth]{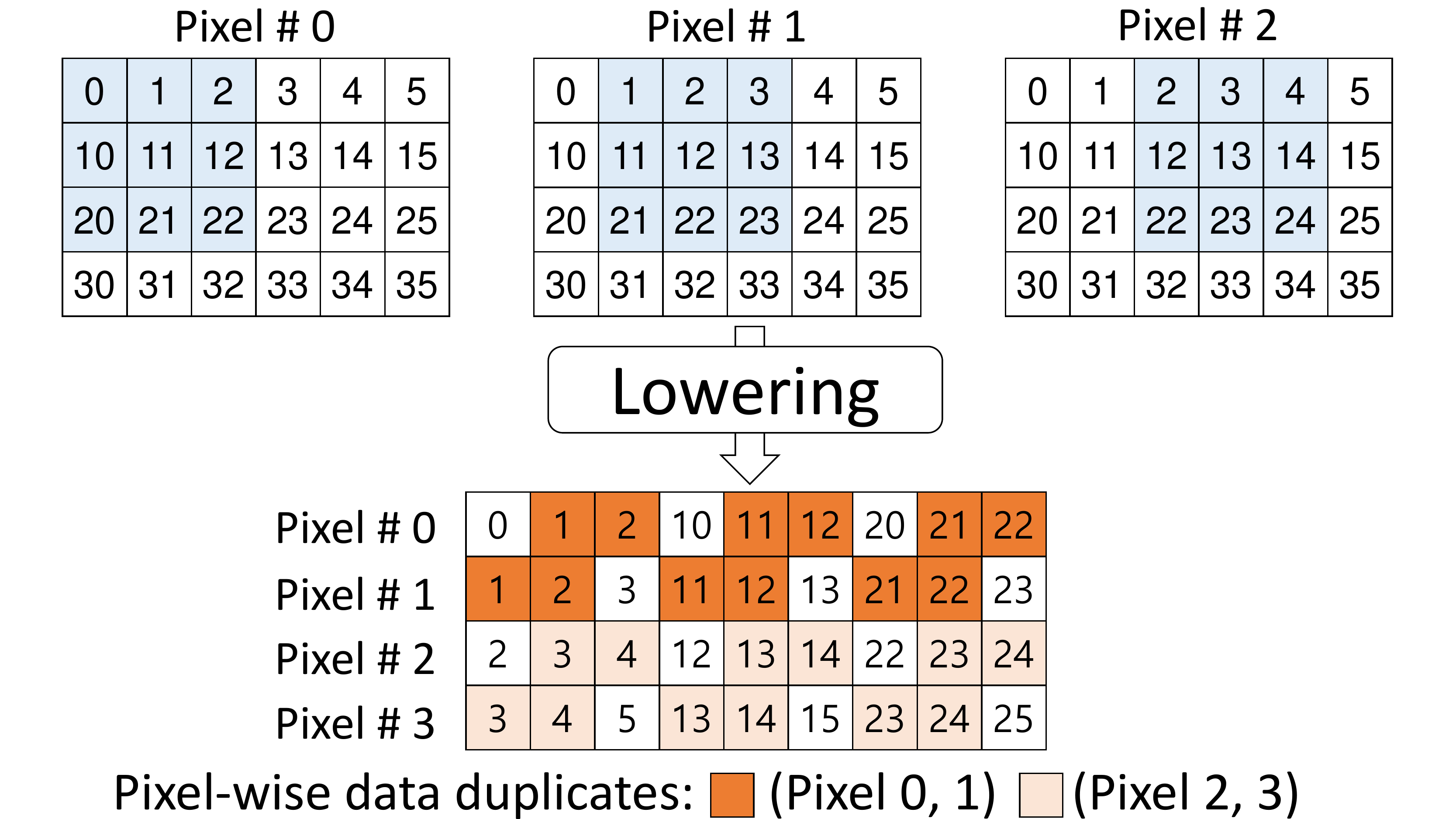}}
\caption{Pixel-wise duplicates generated on the process of lowering.}
\label{lowering}
\end{center}
\vskip -0.3in
\end{figure}

\begin{figure}[t]
\vskip 0.1in
\begin{center}
\centerline{\includegraphics[width=\columnwidth]{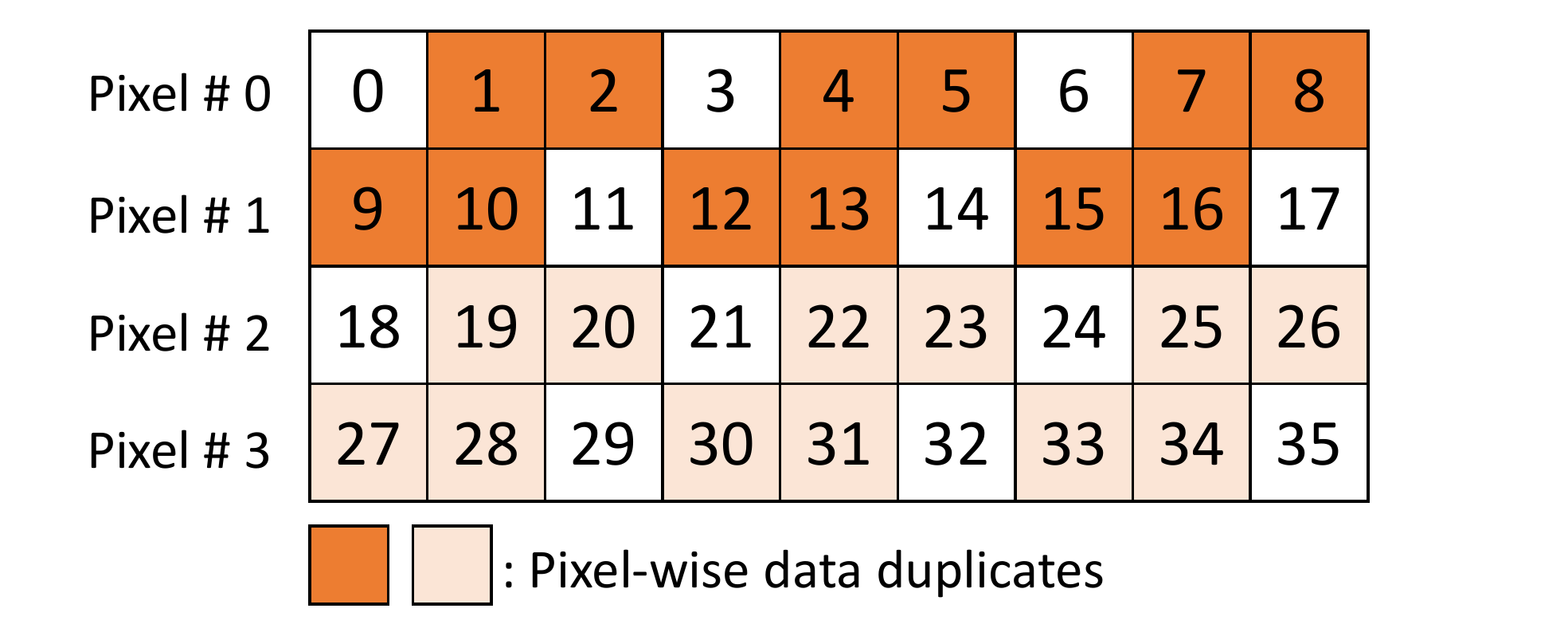}}
\caption{Index view of duplicates and possibility of compiler's static awareness.}
\label{lowering_index_view}
\end{center}
\vskip -0.3in
\end{figure}

It appears that the lowered feature map data contains a considerable amount of data duplicates in Figure~\ref{lowering}. As 3x3 convolution kernel sweeps every pixel on the original feature map, required feature map data for each pixel overlap each other, resulting in pixel-wise data duplicates. The bigger the kernel size the convolution has, the more duplicates the lowered feature map contains. 

Note that the position of pixel-level data duplicates is determined by the algorithm-level convolution configuration such as convolution filter size, filter stride, and feature map size, which can be informed to the compiler. In other words, once a high-level convolution configuration is given, the compilation system can statically predict where the duplicates exist. As shown in Figure~\ref{lowering_index_view}, the compiler already has information on a data tuple with index (1, 2) having the same data with a data tuple with index (9, 10). Likewise, every index with data duplicate can be told and be informed to the code generator. Based on the duplicate information, the code generator can generate code that can reduce the number of memory loads by not fetching data duplicates.

\subsubsection{implementation}

Among duplicates on the feature map, let's say one of the duplicates is the \emph{\textbf{genuine datum}} and the others are \emph{\textbf{duplicate data}}. Also, name the index pointing to the genuine datum as \emph{\textbf{genuine index}} and index pointing to the duplicate datum as \emph{\textbf{duplicate index}}. For example, in Figure~\ref{lowering_index_view}, if index 8 is a genuine index for a datum 22, index 16 and 24 become duplicate indices for the corresponding datum. Then, the duplicate index to genuine index many-to-one-mapping represents the whole duplicates information of the lowered feature map data.

\begin{algorithm}[tb]
    \caption{Duplicate-aware load}
    \label{alg:dup_aware_load}
\begin{algorithmic}[1]
    \State Followings omit weight data load for simplicity:
    \State $Duplicates\_info = Compiler.get\_info()$
    \ForEach{ThreadBlock}
    \ForEach{$dst$ in genuine\_idx}
    \State $f\_shared[dst] = f\_global[src]$
    \EndFor
    \ForEach{ThreadWarp}
    \ForEach{$dst$ in genuine\_idx}
    \State $src = get\_genuine(src)$
    \State $f\_reg[dst] = f\_shared[src]$
    \EndFor
    \ForEach{$f\_idx$}
    \State $f\_idx = get\_genuine(f\_idx)$
    \ForEach{$w\_idx$}
    \State $MMA(f\_reg[f\_idx], w\_reg[w\_idx])$
    \EndFor
    \EndFor
    \EndFor
    \EndFor
\end{algorithmic}
\end{algorithm}

The main concept of the duplicate-aware load is to transform the address space with duplicates into the address space only with genuine data via the duplicates index information. By transforming the memory space itself into data space without any duplicates, every data access cannot structurally generate any data duplicate. This can be done both on GPU shared memory and GPU registers. Algorithm~\ref{alg:dup_aware_load} shows the actual implementation of the duplicate-aware load. On the compile time, the compiler can identify the information about the position of the duplicates with a high-level definition of the convolution. (line 2)  When loading feature map data from global memory into shared memory, (line 5) restrict the data load destination only to genuine index, not allowing any duplicates to exist on the shared memory. (line 4) Note that there are no duplicates on the original global memory.  Likewise, when loading feature map from shared memory to GPU register tile, transfer both source (line 9) and destination index (line 8) to genuine index space. Source index transfer is for the consistency of indexing on the shared memory and destination index transfer is, again, not to allow any duplicates on the register tiles. Lastly, the indices for final MMA computation are set to genuine index space for consistency. (line 13) Note that the address and index control Algorithm~\ref{alg:dup_aware_load} shows is conducted statically by the compiler using convolution configuration on compile time. Thus there is no runtime overhead of the control.

\subsection{Register-level Data Packing}
\label{sec:data_packing}

\subsubsection{data bits required for accumulation}
MMA(Matrix Multiply and Accumulation) execution, as its name suggests, consists of scalar multiplication and accumulation. The execution therefore needs an accumulation register to accumulate values on every vector multiplication. If we assume a 4-bit integer convolution with 128 input channels for accumulation, the accumulation register requires 16 data bits for the maximum possible value ($2^4 * 2^4 * 128 = 2^{15}$). Furthermore, quantized neural networks rarely exhibit such data with excessive magnitude.\cite{pact2018}. However, NVIDIA output tile accumulator sets the accumulator bit width for 4-bit convolution as 32-bit, wasting a great number of bits as zeroes. Numerically, we need about 1 million input convolution channels for accumulation to fully utilize the 32-bit accumulator on 4-bit 3x3 convolution.

\subsubsection{post-convolution data processing}

\begin{figure}[t]
\vskip 0.0in
\begin{center}
\centerline{\includegraphics[width=\columnwidth]{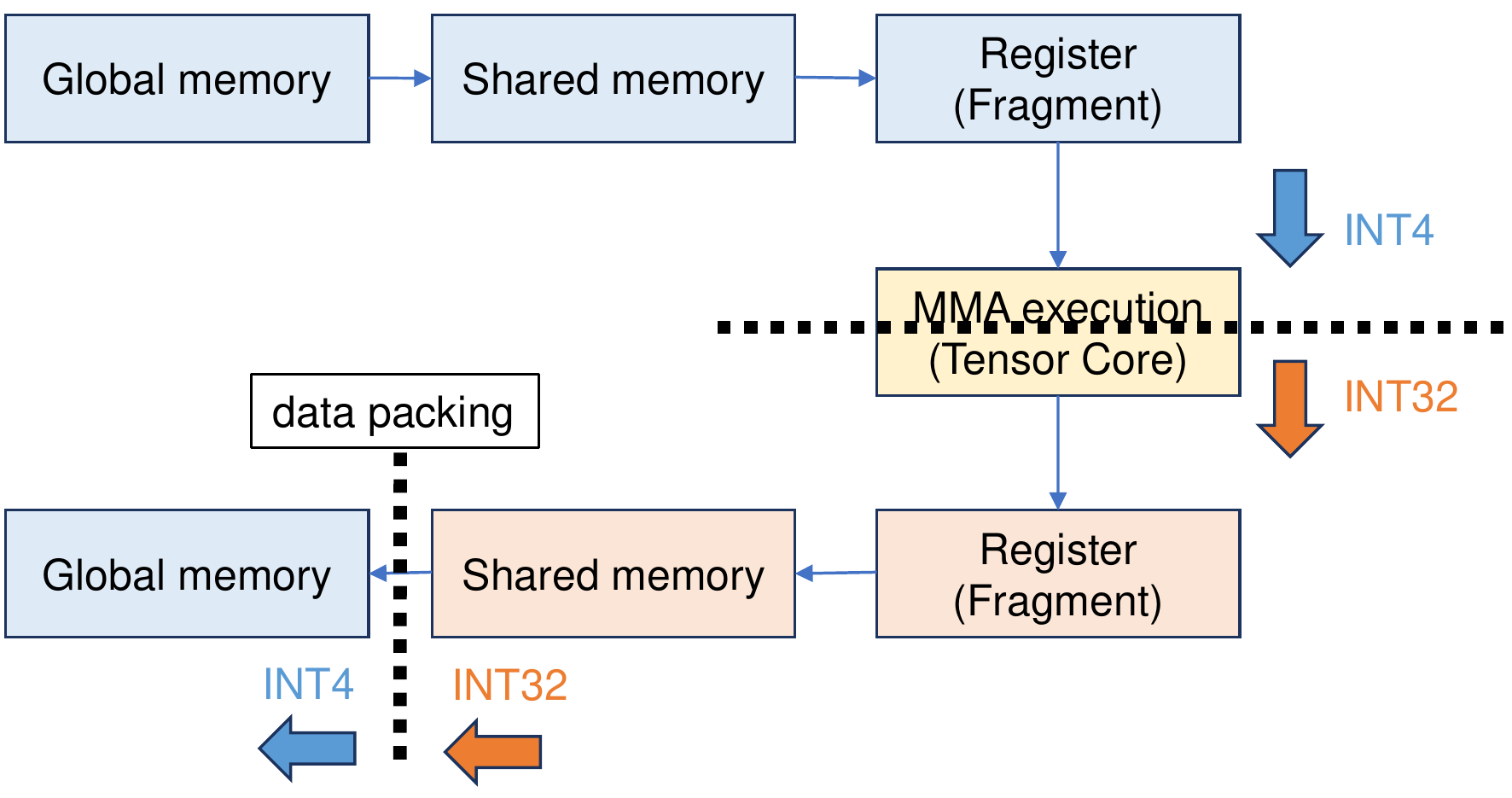}}
\caption{Excessive usage of shared memory before data packing.}
\label{before_data_packing}
\end{center}
\vskip -0.4in
\end{figure}

\begin{figure}[t]
\vskip 0.0in
\begin{center}
\centerline{\includegraphics[width=\columnwidth]{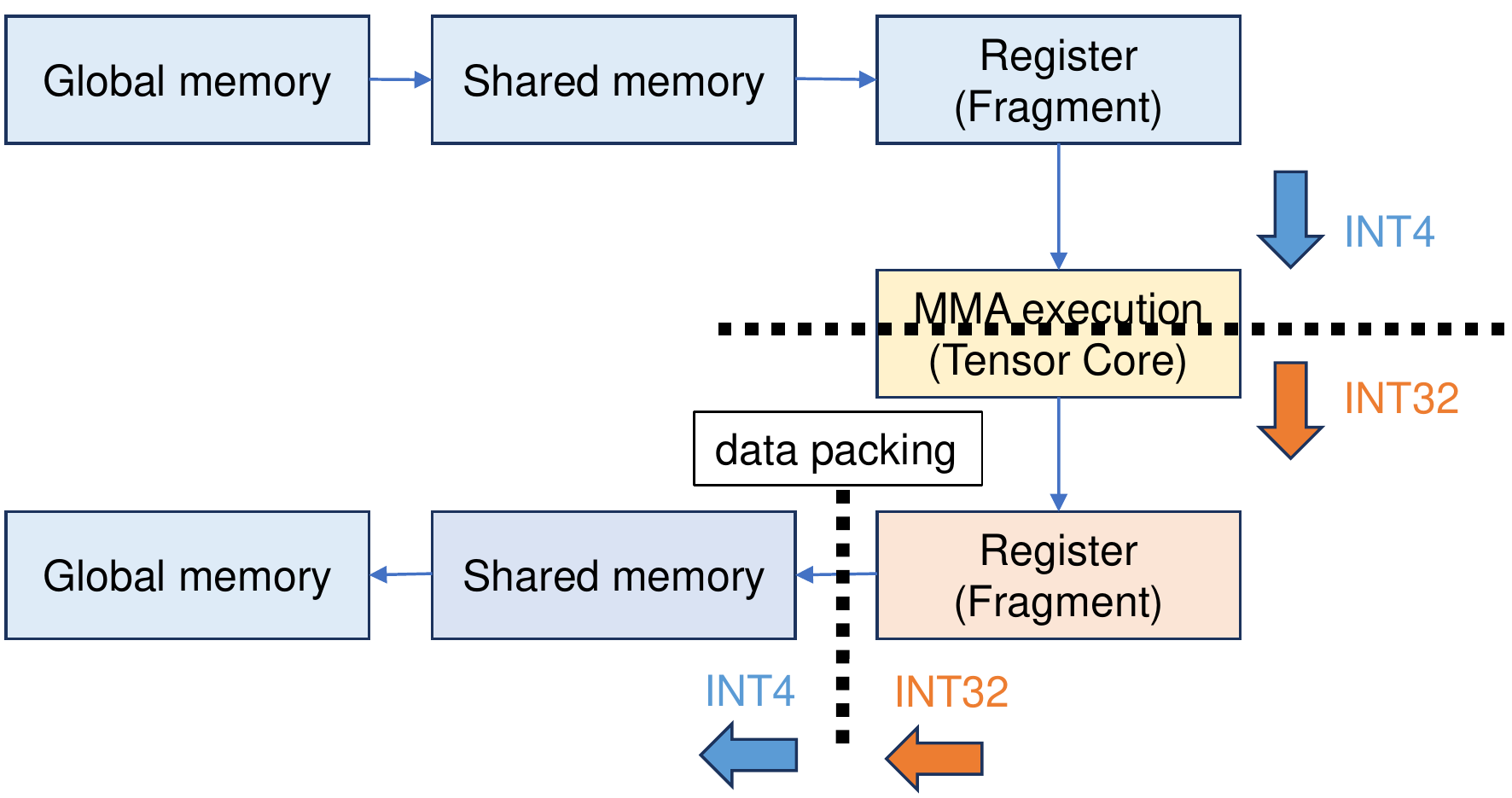}}
\caption{Reduce usage of shared memory after data packing.}
\label{after_data_packing}
\end{center}
\vskip -0.4in
\end{figure}

\begin{figure}[t]
\vskip 0.2in
\begin{center}
\centerline{\includegraphics[width=\columnwidth]{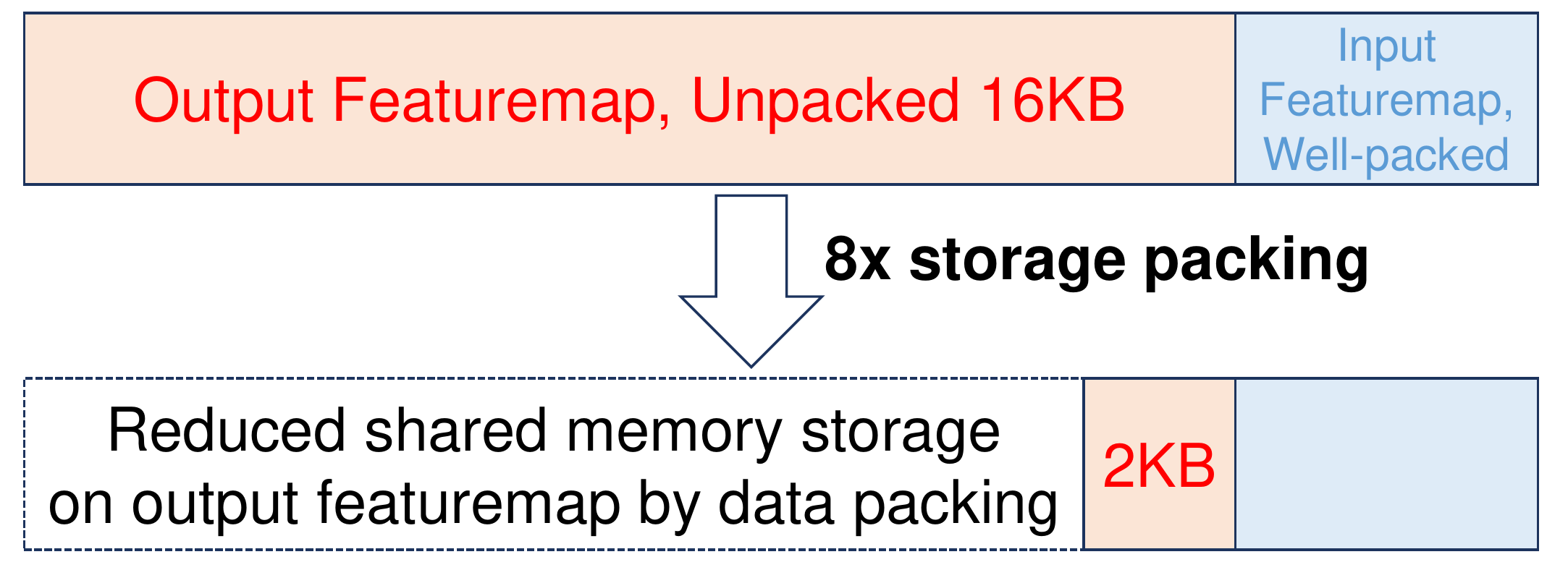}}
\caption{Effective shared memory size reduction after output data packing}
\label{shared_memory_size}
\end{center}
\vskip -0.4in
\end{figure}

On convolutional neural networks, after the convolution computation, remaining epilogue operations such as relu, batch normalization, and bias addition should be computed. We found that the accumulated data with high bit width are stored directly to shared memory for further epilogue processing, wasting a great amount of cache memory bandwidth. After the epilogue processing, the result data are finally clipped to lower bits and packed into 32-bit integer datatype (Figure~\ref{before_data_packing}). It is because lower bit clipping should not be done before all computations including the epilogue have ended. That is, on the other hand, lower bit clipping can be conducted after the epilogue processing is done. If we reorder the position of epilogue processing before storing data to shared memory (Figure~\ref{after_data_packing}), clipping and packing data can be carried out before shared memory store, saving a great amount of GPU cache bandwidth. Furthermore, as the allocated shared memory size for the output feature map is reduced (Figure~\ref{shared_memory_size}), we can allocate more thread blocks on the GPU SM(Streaming Multiprocessors) due to relaxed L1 constraints, reinforcing better parallelism.

\subsubsection{implementation}
\begin{figure}[t]
\vskip 0.0in
\begin{center}
\centerline{\includegraphics[width=\columnwidth]{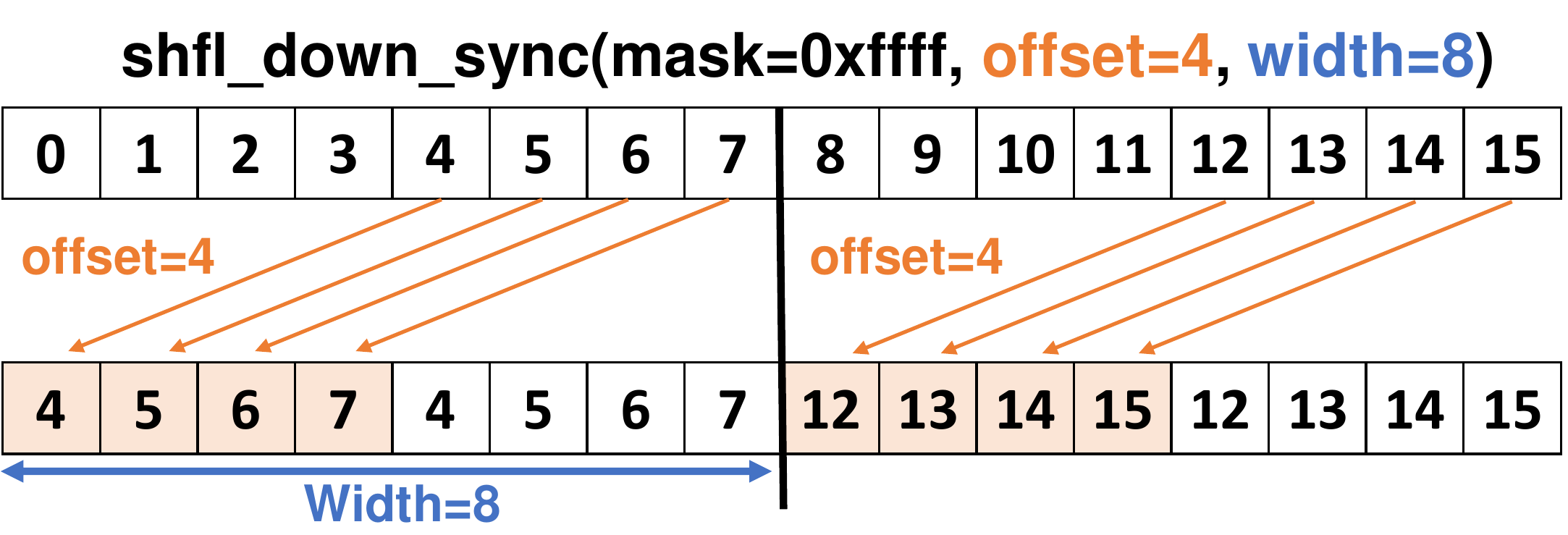}}
\caption{Warp shuffle instrinsic function.}
\label{warp_shuffle}
\end{center}
\vskip -0.4in
\end{figure}

As clipped and well-packed data should be constructed before the shared memory store, the whole data packing process has to be handled on the registers. However, especially for intra-warp data gathering/scattering, a programmer cannot have full control of the registers as a single warp(32 threads) operates as an atomic unit of execution. Instead, a programmer can make use of CUDA's "warp shuffle" intrinsic function, which allows threads to read or write to each other's data registers in a predefined way. Figure~\ref{warp_shuffle} illustrates how a warp shuffle function works. Before a warp shuffle, sixteen threads are holding a single data on each register. Warp shuffle function with offset 4 moves a data of thread 4 to thread 0, thread 5 to thread 1, and so on. Such data movement is conducted on the granularity of warp shuffle width, in this case, 8.

\begin{figure}[t]
\vskip 0.0in
\begin{center}
\centerline{\includegraphics[width=\columnwidth]{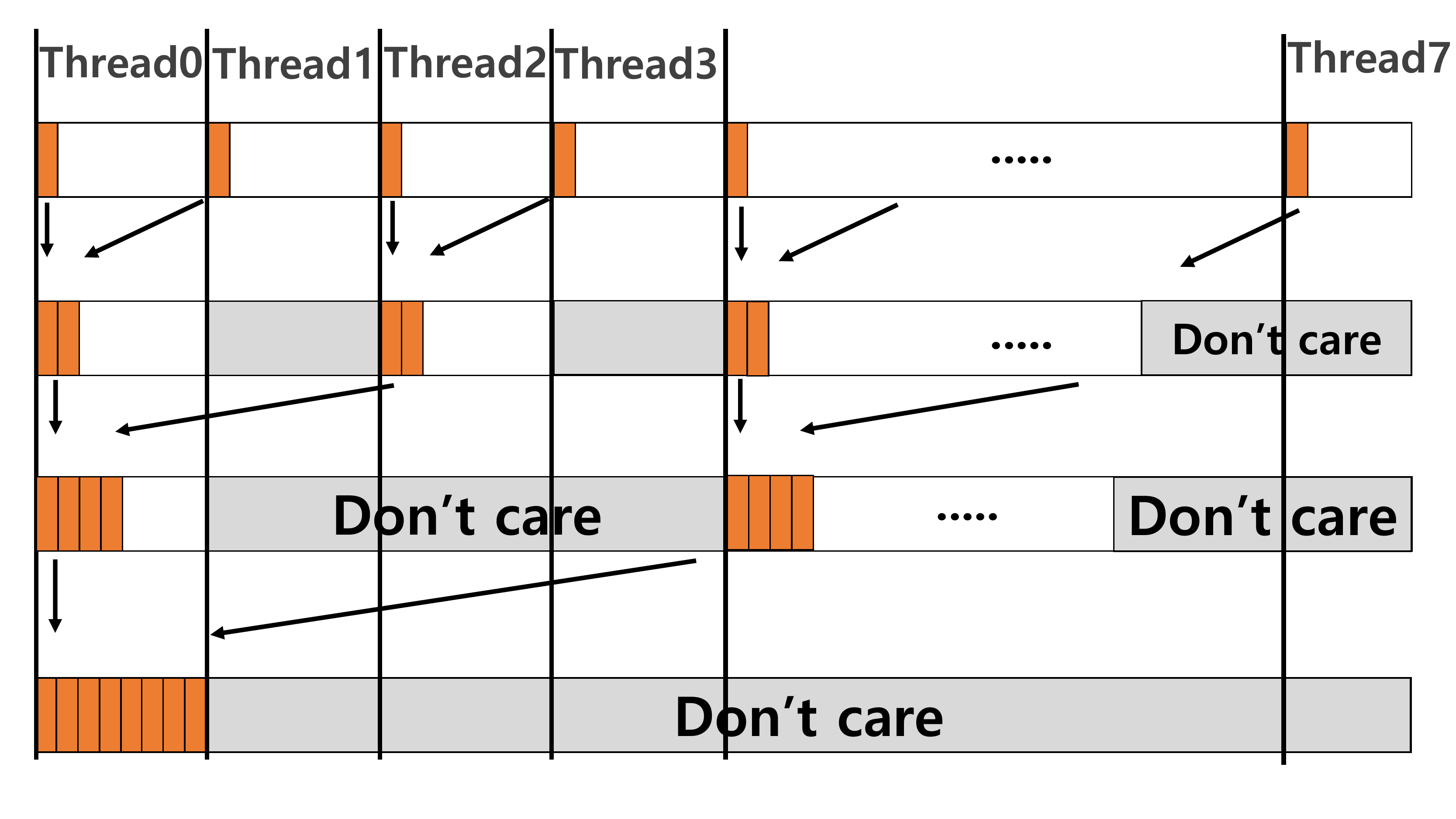}}
\caption{Register level output data packing.}
\label{register-level-packing}
\end{center}
\vskip -0.4in
\end{figure}

\begin{figure}[t]
\vskip 0.2in
\begin{center}
\centerline{\includegraphics[width=\columnwidth]{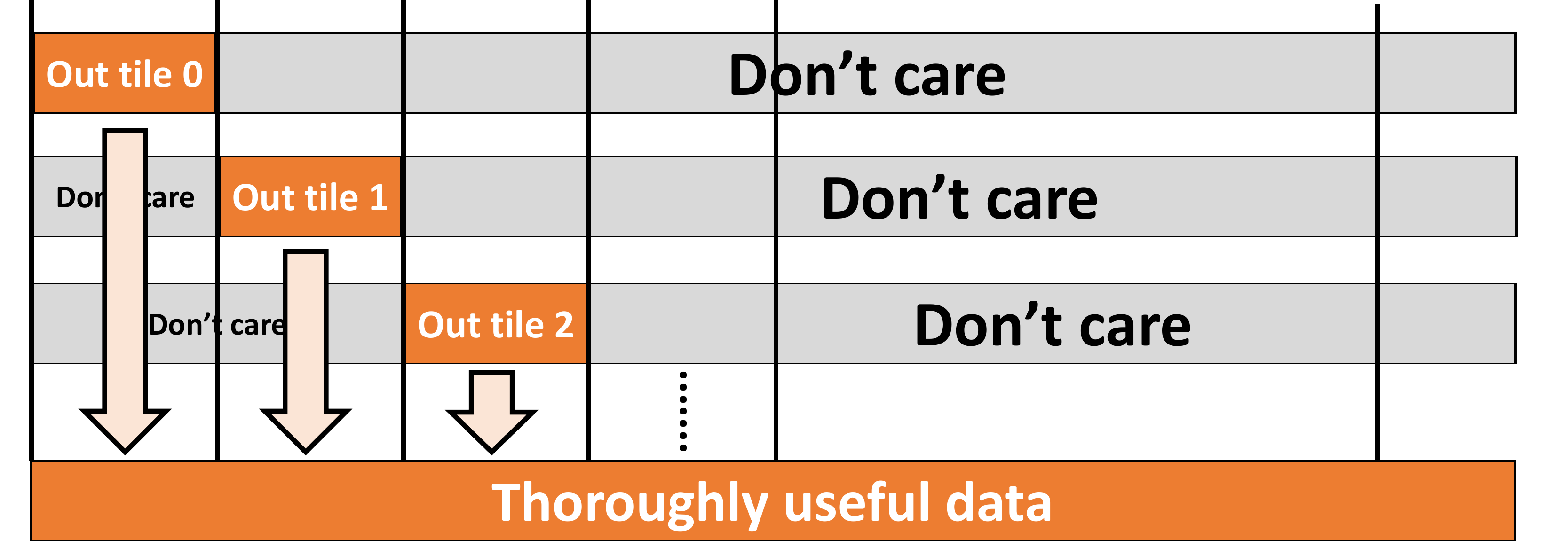}}
\caption{Filling in useful data into the registers after packing.}
\label{packing_postprocess}
\end{center}
\vskip -0.2in
\end{figure}

Using warp shuffling, we can implement intra-warp, register-level data packing on a GPU. Figure~\ref{register-level-packing} shows the actual CUDA implementation of 4-bit integer packing inside a 32-bit integer data register. As you can see in Figure~\ref{register-level-packing}, each 4-bit integer datum is gathered and packed inside a single warp in an iterative process. After the final step of the iterative process, only a single thread among the eight threads holds the useful data which may result in useless store requests even with data packing. To avoid this inefficiency, gather the other output register tiles into registers with don't care values so that all store requests from all registers can be meaningful. (Figure~\ref{packing_postprocess})

\subsection{Data Layout Aware Register Manipulation}
\subsubsection{coalesced memory access on GPU}
For the best practice of GPGPU programming, all global memory access should be \emph{\textbf{coalesced}}, which means that addresses for memory access should not diverge inside a single warp. Otherwise, useless data transactions may occur. This is because the atomic data access unit is 32-byte in modern GPU architecture so consecutive huge accesses perform better than divergent small accesses.

\begin{figure}[t]
\vskip 0.0in
\begin{center}
\centerline{\includegraphics[width=\columnwidth]{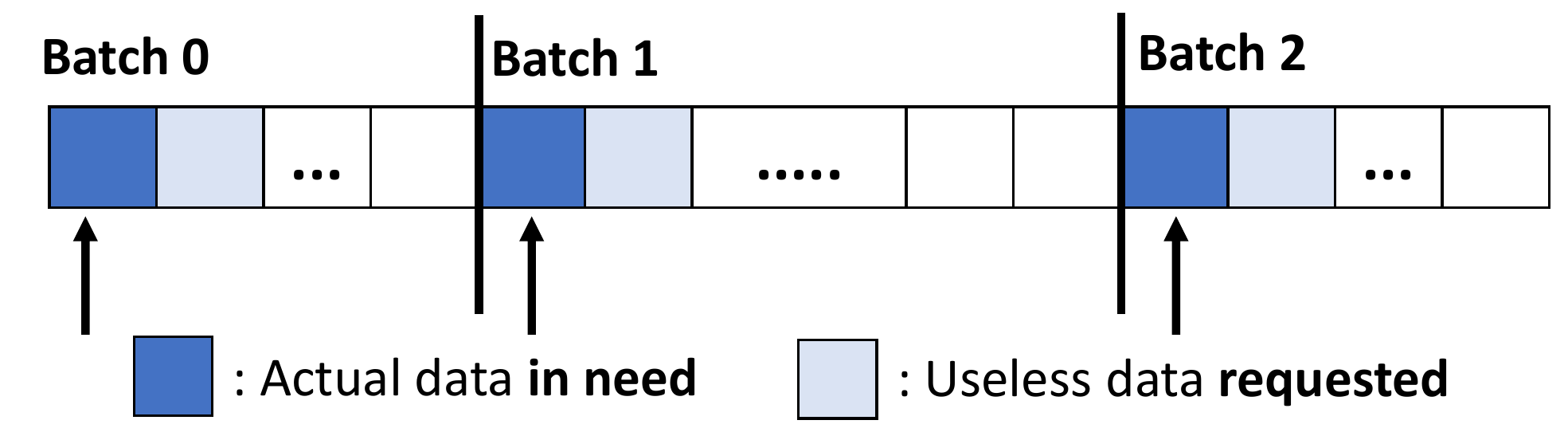}}
\caption{Uncoalesced access. Each box represents 16-byte data.}
\label{uncoalesed_access}
\end{center}
\vskip -0.4in
\end{figure}

Meanwhile, for Tensor Core operation, the input feature map tensor should be split into atomic WMMA register tile size. That is, NHWC data layout should be reshaped into NHWCnc layout where `n' stands for the input feature map batch size as a row size of WMMA register tile and `c' stands for the input feature map channel size as a column size of WMMA tile. We found that such data layout reshape results in uncoalesced memory access because the size of the lowest channel dimension `c' is 16-bytes-wide because of GPU hardware intrinsic size of CUDA WMMA operation, which does not satisfy the atomic 32-bytes load of the GPU memory system. Moreover, such 16-bytes-wide memory accesses are diverged among batches, resulting in the failure of consecutive load (Figure~\ref{uncoalesed_access}).

\subsubsection{Consideration on the Data layout}
To avoid this uncoalesced memory access, we suggest global memory layout be stored as NHWCnc layout rather than NHWC when using NVIDIA Tensor Cores. Otherwise, all GPU kernels will suffer from uncoalesced memory access. However, additional effort is required to consistently maintain the same data layout. The number of output feature map channels depends on the number of filters, not the number of input feature map channels. So we cannot guarantee the last channel dimension(`c') of the output feature map would be the same as the input feature map after the convolution execution. once we are done with register-level data packing which section~\ref{sec:data_packing} suggests, the last channel dimension gets even more reformed.

Nevertheless, the `n' and `c' dimensions are originated from the warp level MMA operation, which means that the sub-tensor with such two dimensions is intentionally covered with registers inside a single warp. Therefore, for consistent data layout maintenance, only an additional single warp shuffle intrinsic function is required.

\subsection{Diversity-Aware Schedule Search}

\begin{figure}[t]
\begin{center}
\centerline{\includegraphics[width=\columnwidth]{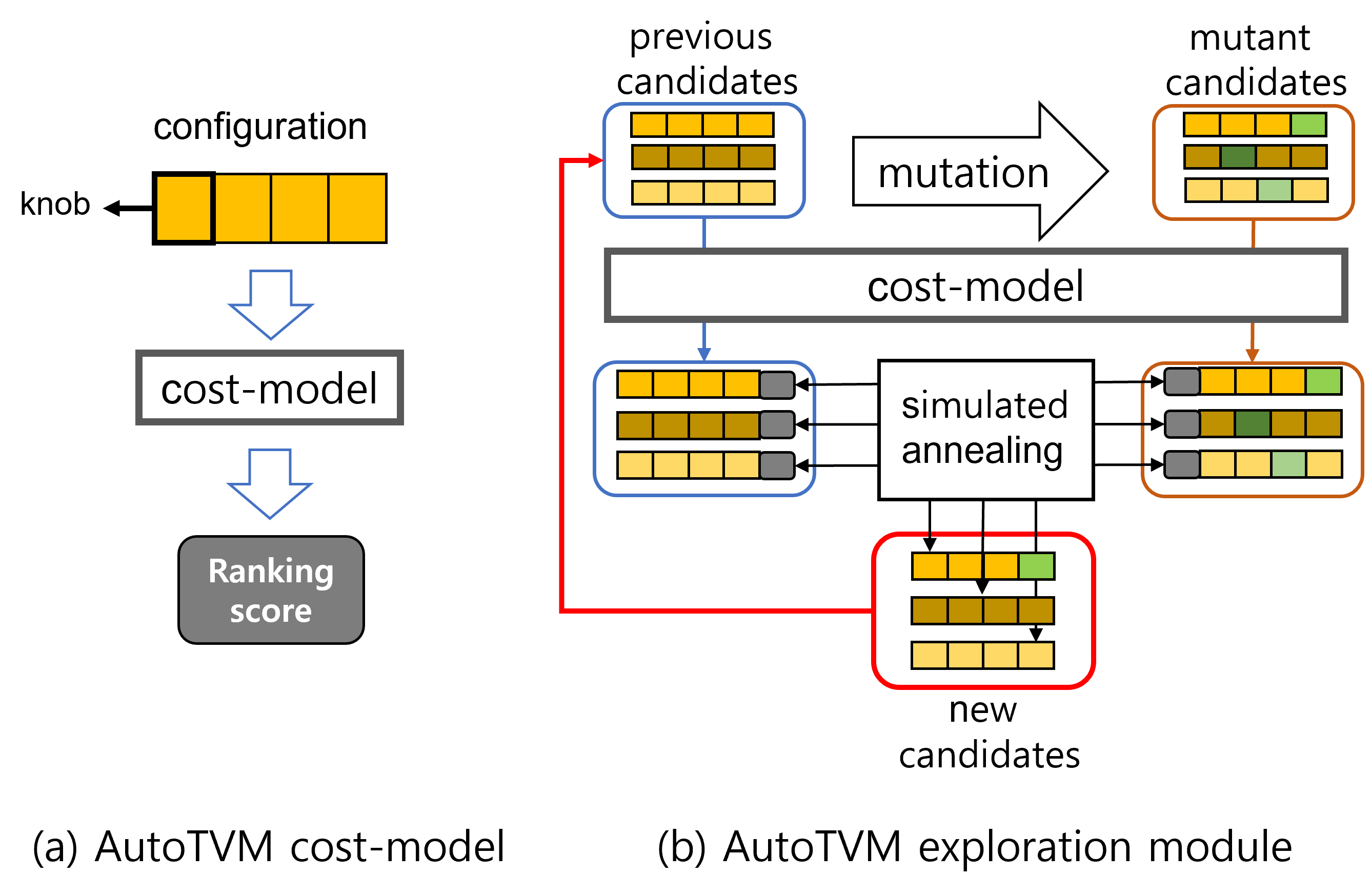}}
\caption{AutoTVM cost-model and exploration module}
\label{exploration module}
\end{center}
\vskip -0.4in
\end{figure}

\begin{figure}[t]
\begin{center}
\centerline{\includegraphics[width=\columnwidth]{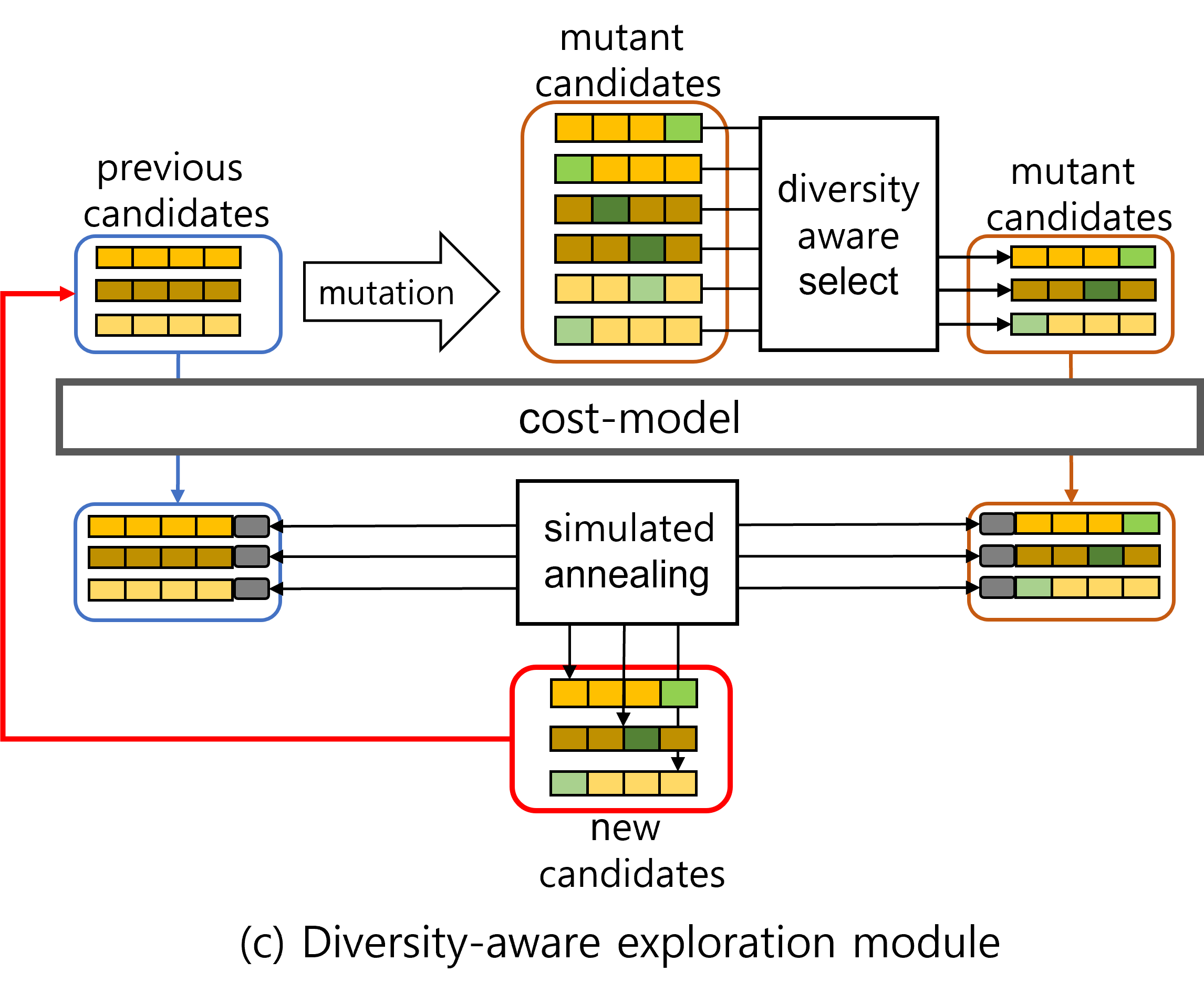}}
\caption{Exploration module with diversity-aware selection}
\label{exploration module}
\end{center}
\vskip -0.4in
\end{figure}

Existing AutoTVM suggests a simulated-annealing-algorithm-based exploration module with statistical cost-model to efficiently search for optimal configuration in large exhaustive search space (Figure 12).

Statistical cost-model allows exploration module to compare which configuration’s low-level code is faster without real hardware measurement. The cost model is trained by {(configuration, runtime)} dataset with ranking loss objective function. AutoTVM can evaluate many configurations in a short time using this cost model.

The exploration module picks candidates which are expected to optimal configuration using a simulated annealing algorithm and cost-model (Figure 12-(b)). The exploration module repeats the selection a set number of times.
To search candidates efficiently in a large search space, the exploration module mutate one random knob of previous candidates to explore better performance candidates. the exploration module uses a simulated annealing algorithm with the cost-model score as an energy function. The algorithm compares previous candidates and mutant candidates to pick better candidates repeatedly, finally selecting the top-performing batch of candidates to run on real hardware. The result of candidates' performance data is used to update the cost-model. 

The weakness of AutoTVM is Exploration module picks too many similar candidates that have similar performance and this does not help to improve the quality of the cost model. 
At the beginning of the AutoTVM search, the cost-model is only trained with a small number of randomly selected configurations. This cost model is hard to predict diverse configurations accurately. Inside the search space, not all knobs of configuration are critical to performance. Therefore, there are many similar configurations that have similar performance in the search space. The cost model overestimates many configurations similar to the previous best and underestimates many candidates who have not learned it. It causes a simulated annealing algorithm to pick candidates which are too similar to previous candidates and also have similar performance. These new candidates are not useful for improving the cost model quality because similar candidates are already trained.

We propose an improved exploration module that considers diversity-aware search to improve the training efficiency of the cost model (Figure 13).
The proposed exploration module creates two mutant candidates from one previous candidate. Then select half of the entire mutant candidates considering the configuration diversity. The selected mutant candidates compete with previous candidates, improving the quality of the competition. The proposed diversity-aware selection has a positive effects both on the diversity of candidates pool and diversity exploration through mutations. A slight tuning performance improvement was obtained through this method.


\section{Experiments}

\subsection{Experimental Settings}

We performed AutoTVM searches to find the best configuration for the 3x3 spatial feature extraction convolution of each stage of ResNet50. The search space consists of the following six knobs: BLK-ROW-WARPS, BLK-COL-WARPS, WARP-ROW-TILES, WARP-COL-TILES, CHUNK, and REORDER-INNER. BLK-ROW-WARPS and BLK-COL-WARPS refer to the number of warps for M and N dimensions in a single thread block, and similarly, WARP-ROW-TILES and WARP-COL-TILES refer to the number of WMMA tiles for M, N dimensions in a single warp. CHUNK is the loop split factor for input channel accumulation, and REORDER-INNER determines the order between the outer loop of the input channel and the kernel height loop. We searched the best configuration through 500 trials for each convolution, using AutoTVM on the NVIDIA T4 GPU.

Most simulated annealing algorithm settings are the same as previous AutoTVM default settings. The number of iterations of simulated annealing is 500 and stops iteration if the optimal set does not change in 50 rounds. We use the cost-model prediction score as an energy function of the simulated annealing algorithm. The temperature of the algorithm starts from 1 and cooling down 0.002 every iteration. The simulated annealing algorithm picks 128 candidates every iteration and passes them to the next iteration.
At the last Exploration module pick, top-31 configurations from the candidates and one random configuration are added, and those 32 configurations are measured on real hardware. The exploration module only picks candidates that have not been measured before. If there are less than 31 new candidates, randomly generated configurations fill in the rest.

\subsection{Schedule Search on ResNet50}

\begin{table}[tb]
    \centering
    \begin{tabular}{ccccc} \toprule
    Stage                       & 2 & 3 & 4 & 5  \\ \hline
    OPs                       & \multicolumn{4}{c}{1849688064}   \\ \hline
    Baseline (us) & 196.06 & 180.96 & 203.62 & 198.62 \\
    Exhaustive (us) & 50.78 & 51.42 & 57.18 & 86.37 \\ 
    Searched (us) & 50.98 & 50.46 & 55.58 & 70.98 \\ \hline
    Speed-up & 3.85x & 3.59x & 3.66x & 2.80x \\ \bottomrule
    \end{tabular}
    \caption{Performance of 3x3 convolutions in ResNet50 executed with the searched configurations.}
    \label{tab:my_label}
\end{table}

Target convolutions are 3x3 convolutions of each residual block in ResNet50. Stage means a section in which residual blocks of the exact specification are repeated. For instance, stage2 means a residual block with a feature map size of 56x56. And in the next stage, the feature map size is halved, and the channel depth is doubled; therefore, the number of operations remains the same. We used two methods for the search: AutoTVM and exhaustive manual search. As a result, automatic-searched performance is faster or similar because AutoTVM measures many configurations faster than manual search. Baseline is the performance of TVM implementation of GitHub's main branch. This result was also evaluated by finding the optimal configuration with AutoTVM. As shown in Table 1, our searched configurations achieve 2.8x to 3.85x speed-up compared to the baseline for all convolutions.

\subsection{Diversity-Aware Search}

\begin{figure}[t]
\vskip 0.2in
\begin{center}
\centerline{\includegraphics[width=\columnwidth]{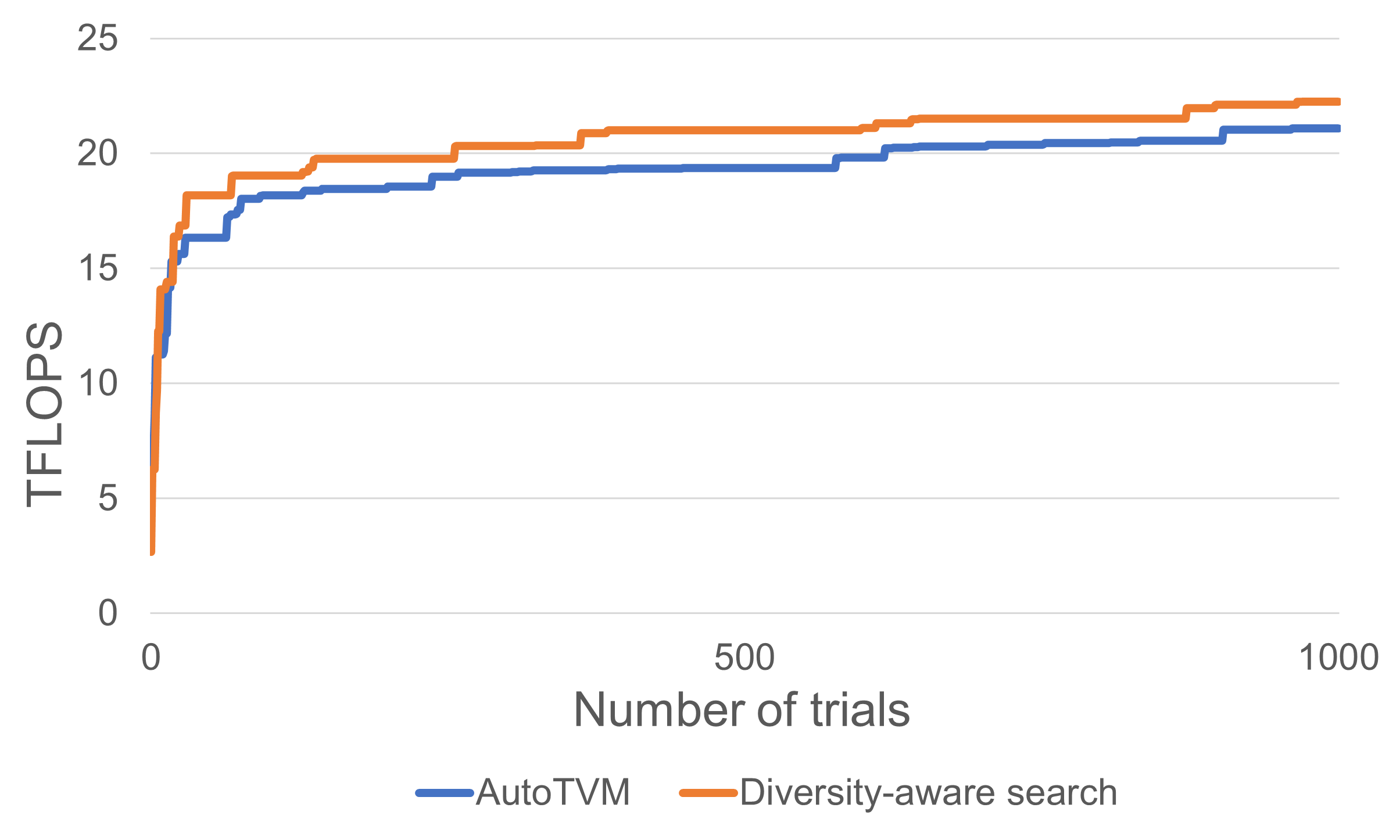}}
\caption{Impact of Diversity-Aware search}

\label{exploration module}
\end{center}
\vskip -0.2in
\end{figure}

Due to the implementation issue, here we conducted the experiments with the search space of the original AutoTVM.
We evaluated the proposed diversity-aware exploration module with AutoTVM's search space. Target convolution is 3x3 convolution of ResNet50 stage 2. Figure 14 shows the diversity-aware search method finds better performance configuration in the same trial.

\subsection{Ablation Study}

\begin{figure}[t]
\begin{center}
\centerline{\includegraphics[width=\columnwidth]{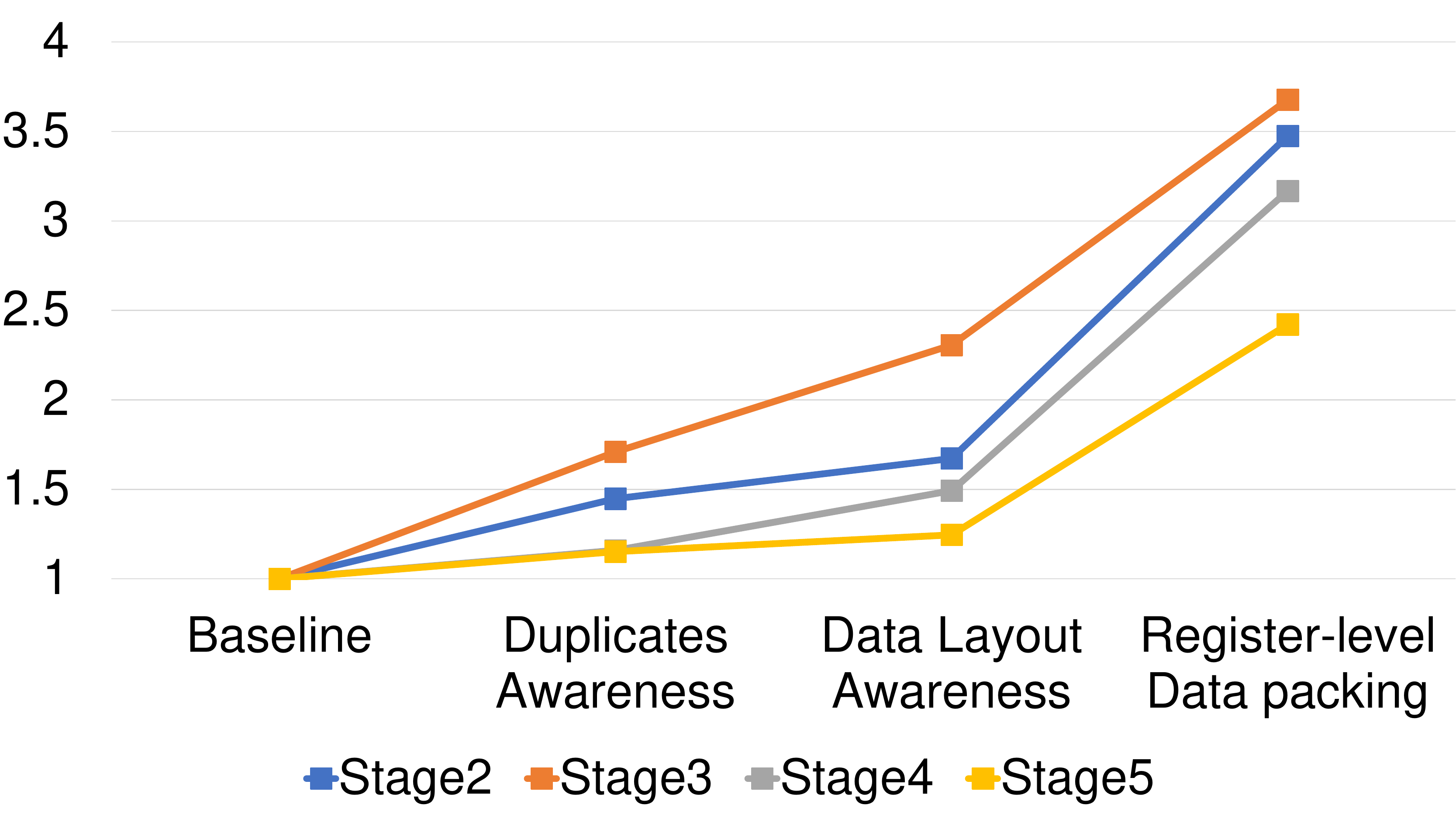}}
\caption{Accumulated Speedup}
\label{accumulated_speedup}
\end{center}
\vskip -0.3in
\end{figure}

\begin{figure}[t]
\begin{center}
\centerline{\includegraphics[width=\columnwidth]{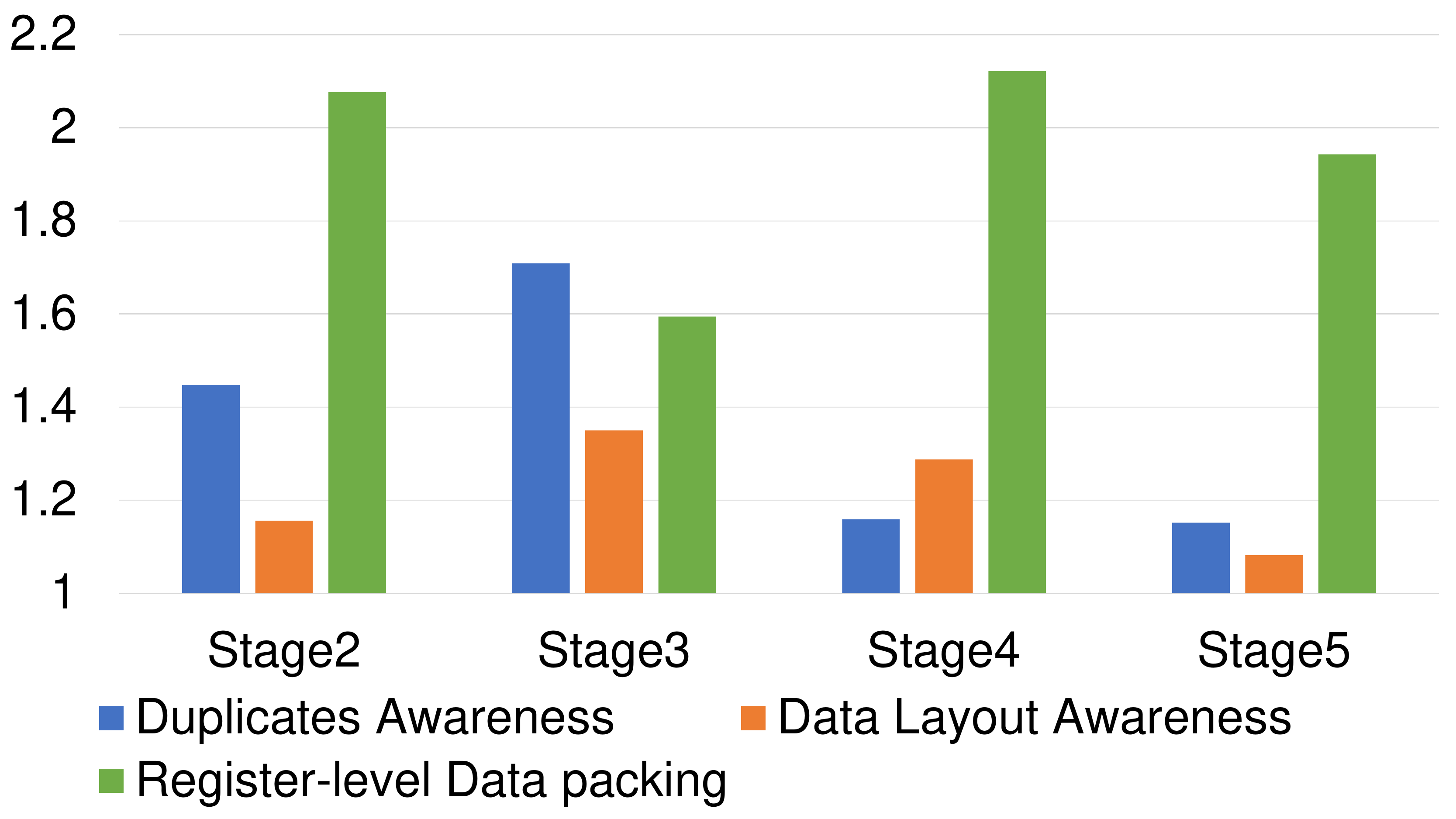}}
\caption{Marginal Speedup}
\label{marginal_speedup}
\end{center}
\vskip -0.4in
\end{figure}

Figure~\ref{accumulated_speedup} depicts accumulated speedup as gradually applying our improvements from the baseline on each convolution. The baseline on each convolution selects the execution schedule with fairly effective performance. Similar to Table~\ref{tab:my_label}, the figure shows that convolution with a larger height \& width size performs better than one with a smaller size. 

While, Figure~\ref{marginal_speedup} presents marginal speedup of each improvement, grouped by types of the convolution. Unlike register-level data packing, which is adequately effective for all convolutions, duplicates awareness does not comparatively perform well on the convolution with smaller width \& height and larger channels \& filters. That is because a massive number of channels obstructs execution schedule to cover a sufficient number of width in a single thread block or warp due to the quality of parallelization. Such scheduling is unfavorable to width-level duplicates awareness which benefits from a multiple width cover in a single execution environment.

\section{Related Work}

There have been numerous studies to accomplish efficient Tensor Core computations. Some studies \cite{duplo2020}\cite{sparsetc2019} tackle this challenge by providing microarchitectural supports to remove redundant memory accesses \cite{duplo2020} or to handle sparsity in neural networks with indexing in the register file \cite{sparsetc2019}. \cite{sparsetc2019}\cite{spgemm2020}\cite{sparsity2021} have also suggested efficient implementation of sparse operations that can hardly obtain performance gain on GPUs due to their irregularity of data layout. Since no algorithm other than GEMM utilizes Tensor Cores, optimizations for non-GEMM algorithms such as reduction and scan have been investigated as well \cite{reduction2020}\cite{scan2019}. Although Tensor Cores can boost the performance of MMA operations by an order of magnitude, optimizing for them is very challenging due to the fixed WMMA size and complicated memory hierarchy. Furthermore, experts have to spend tens to hundreds of hours to manually tune their kernel implementations on GPUs. In order to solve this problem, \cite{autogen2020} has developed automatic kernel generation using polyhedral techniques and implemented an efficient matrix multiplication kernel using the $mma.m8n8k4$ PTX instruction on a Volta GPU.

While various algorithms have been designed for Tensor Cores, there are limited researches on the optimization approaches to utilize reduced precision MMA on Tensor Cores. \cite{winograd2021}\cite{fft2018}\cite{hpmm2020} use reduced precision to achieve faster speed and energy saving for their algorithms. However, only half-precision is used in these works and lower bit-precision types provided by Tensor Cores such as INT8 or INT4 are not investigated. \cite{hawqv32021} proposes to use uniform and mixed-precision INT4/8 quantization for integer-only inference without any floating point operations. It also solves an integer linear programming problem to find the best bit-precision setting. \cite{apnn2021} introduces a framework to support arbitrary precision neural networks with INT1 compute primitives on Tensor Core.

Tensor programs can have numbers of implementations that are logically equivalent but differ significantly in performance because there can be various optimization strategies considering hardware factors such as data locality and pipelining. In order to efficiently explore the wide search space, AutoTVM \cite{autotvm2018} presents a statistical cost model and accelerate the process using transfer learning. However, writing templates used for AutoTVM requires a huge effort, so Ansor \cite{ansor2020} generates the search space automatically without manually developed templates and tune a program using evolutionary search. There also exists an approach for automatically learning a performance model for Tensor Processing Unit (TPU)\cite{tpumodel2021}.


\section{Conclusion}

In this work, we propose an automatic scheduling method of reduced-precision MMA for convolution operation. In this method, we devise a search space that explores the binding options for maximizing utilization of reduced-precision instructions of Tensor Cores. The search space also includes register-level data packing and layout optimization options, which is particularly beneficial for reducing the overhead of handling reduced-precision data. We further propose a search algorithm by learning from the distinctive candidates. Finally, this reduced-precision MMA optimization method is evaluated on convolution operations of popular neural networks with superior performance compared to the state-of-the-art.

\bibliography{main}

\begin{thebibliography}{22}
\providecommand{\natexlab}[1]{#1}
\providecommand{\url}[1]{\texttt{#1}}
\expandafter\ifx\csname urlstyle\endcsname\relax
  \providecommand{\doi}[1]{doi: #1}\else
  \providecommand{\doi}{doi: \begingroup \urlstyle{rm}\Url}\fi

\bibitem[Bhaskaracharya et~al.(2020)Bhaskaracharya, Demouth, and
  Grover]{autogen2020}
Bhaskaracharya, S.~G., Demouth, J., and Grover, V.
\newblock Automatic kernel generation for volta tensor cores.
\newblock \emph{arXiv preprint arXiv:2006.12645}, 2020.

\bibitem[Carrasco et~al.(2018)Carrasco, Vega, and
  Navarro]{carrasco2018analyzing}
Carrasco, R., Vega, R., and Navarro, C.~A.
\newblock Analyzing gpu tensor core potential for fast reductions.
\newblock In \emph{2018 37th International Conference of the Chilean Computer
  Science Society (SCCC)}, pp.\  1--6. IEEE, 2018.

\bibitem[Chen et~al.(2018)Chen, Zheng, Yan, Jiang, Moreau, Ceze, Guestrin, and
  Krishnamurthy]{autotvm2018}
Chen, T., Zheng, L., Yan, E., Jiang, Z., Moreau, T., Ceze, L., Guestrin, C.,
  and Krishnamurthy, A.
\newblock Learning to optimize tensor programs.
\newblock NIPS'18, pp.\  3393–3404, 2018.

\bibitem[Chen et~al.(2021)Chen, Qu, Liu, Ding, and Xie]{sparsity2021}
Chen, Z., Qu, Z., Liu, L., Ding, Y., and Xie, Y.
\newblock Efficient tensor core-based gpu kernels for structured sparsity under
  reduced precision.
\newblock In \emph{Proceedings of the International Conference for High
  Performance Computing, Networking, Storage and Analysis}, SC '21, 2021.

\bibitem[Choi et~al.(2018)Choi, Wang, Venkataramani, Chuang, Srinivasan, and
  Gopalakrishnan]{pact2018}
Choi, J., Wang, Z., Venkataramani, S., Chuang, P.~I., Srinivasan, V., and
  Gopalakrishnan, K.
\newblock {PACT:} parameterized clipping activation for quantized neural
  networks.
\newblock \emph{CoRR}, abs/1805.06085, 2018.

\bibitem[Dakkak et~al.(2019)Dakkak, Li, Xiong, Gelado, and Hwu]{scan2019}
Dakkak, A., Li, C., Xiong, J., Gelado, I., and Hwu, W.-m.
\newblock Accelerating reduction and scan using tensor core units.
\newblock In \emph{Proceedings of the ACM International Conference on
  Supercomputing}, pp.\  46--57, 2019.

\bibitem[Dosovitskiy et~al.(2020)Dosovitskiy, Beyer, Kolesnikov, Weissenborn,
  Zhai, Unterthiner, Dehghani, Minderer, Heigold, Gelly,
  et~al.]{dosovitskiy2020image}
Dosovitskiy, A., Beyer, L., Kolesnikov, A., Weissenborn, D., Zhai, X.,
  Unterthiner, T., Dehghani, M., Minderer, M., Heigold, G., Gelly, S., et~al.
\newblock An image is worth 16x16 words: Transformers for image recognition at
  scale.
\newblock \emph{arXiv preprint arXiv:2010.11929}, 2020.

\bibitem[Feng et~al.(2021)Feng, Wang, Geng, Li, and Ding]{apnn2021}
Feng, B., Wang, Y., Geng, T., Li, A., and Ding, Y.
\newblock Apnn-tc: Accelerating arbitrary precision neural networks on ampere
  gpu tensor cores.
\newblock In \emph{Proceedings of the International Conference for High
  Performance Computing, Networking, Storage and Analysis}, pp.\  1--13, 2021.

\bibitem[He et~al.(2016)He, Zhang, Ren, and Sun]{he2016deep}
He, K., Zhang, X., Ren, S., and Sun, J.
\newblock Deep residual learning for image recognition.
\newblock In \emph{Proceedings of the IEEE conference on computer vision and
  pattern recognition}, pp.\  770--778, 2016.

\bibitem[Kaufman et~al.(2021)Kaufman, Phothilimthana, Zhou, Mendis, Roy, Sabne,
  and Burrows]{tpumodel2021}
Kaufman, S., Phothilimthana, P., Zhou, Y., Mendis, C., Roy, S., Sabne, A., and
  Burrows, M.
\newblock A learned performance model for tensor processing units.
\newblock \emph{Proceedings of Machine Learning and Systems}, 3, 2021.

\bibitem[Kim et~al.(2020)Kim, Ahn, Oh, Kim, Ro, and Song]{duplo2020}
Kim, H., Ahn, S., Oh, Y., Kim, B., Ro, W.~W., and Song, W.~J.
\newblock Duplo: Lifting redundant memory accesses of deep neural networks for
  gpu tensor cores.
\newblock In \emph{2020 53rd Annual IEEE/ACM International Symposium on
  Microarchitecture (MICRO)}, pp.\  725--737. IEEE, 2020.

\bibitem[Krizhevsky et~al.(2012{\natexlab{a}})Krizhevsky, Sutskever, and
  Hinton]{alexnet2021}
Krizhevsky, A., Sutskever, I., and Hinton, G.~E.
\newblock Imagenet classification with deep convolutional neural networks.
\newblock \emph{Advances in neural information processing systems},
  25:\penalty0 1097--1105, 2012{\natexlab{a}}.

\bibitem[Krizhevsky et~al.(2012{\natexlab{b}})Krizhevsky, Sutskever, and
  Hinton]{krizhevsky2012imagenet}
Krizhevsky, A., Sutskever, I., and Hinton, G.~E.
\newblock Imagenet classification with deep convolutional neural networks.
\newblock \emph{Advances in neural information processing systems},
  25:\penalty0 1097--1105, 2012{\natexlab{b}}.

\bibitem[Liu et~al.(2021)Liu, Yang, and Lai]{winograd2021}
Liu, J., Yang, D., and Lai, J.
\newblock Optimizing winograd-based convolution with tensor cores.
\newblock In \emph{50th International Conference on Parallel Processing}, pp.\
  1--10, 2021.

\bibitem[Navarro et~al.(2020)Navarro, Carrasco, Barrientos, Riquelme, and
  Vega]{reduction2020}
Navarro, C.~A., Carrasco, R., Barrientos, R.~J., Riquelme, J.~A., and Vega, R.
\newblock Gpu tensor cores for fast arithmetic reductions.
\newblock \emph{IEEE Transactions on Parallel and Distributed Systems},
  32\penalty0 (1):\penalty0 72--84, 2020.

\bibitem[Sorna et~al.(2018)Sorna, Cheng, D'azevedo, Won, and Tomov]{fft2018}
Sorna, A., Cheng, X., D'azevedo, E., Won, K., and Tomov, S.
\newblock Optimizing the fast fourier transform using mixed precision on tensor
  core hardware.
\newblock In \emph{2018 IEEE 25th International Conference on High Performance
  Computing Workshops (HiPCW)}, pp.\  3--7. IEEE, 2018.

\bibitem[Szegedy et~al.(2016)Szegedy, Vanhoucke, Ioffe, Shlens, and
  Wojna]{szegedy2016rethinking}
Szegedy, C., Vanhoucke, V., Ioffe, S., Shlens, J., and Wojna, Z.
\newblock Rethinking the inception architecture for computer vision.
\newblock In \emph{Proceedings of the IEEE conference on computer vision and
  pattern recognition}, pp.\  2818--2826, 2016.

\bibitem[Yan et~al.(2020)Yan, Wang, and Chu]{hpmm2020}
Yan, D., Wang, W., and Chu, X.
\newblock Demystifying tensor cores to optimize half-precision matrix multiply.
\newblock In \emph{2020 IEEE International Parallel and Distributed Processing
  Symposium (IPDPS)}, pp.\  634--643. IEEE, 2020.

\bibitem[Yao et~al.(2021)Yao, Dong, Zheng, Gholami, Yu, Tan, Wang, Huang, Wang,
  Mahoney, et~al.]{hawqv32021}
Yao, Z., Dong, Z., Zheng, Z., Gholami, A., Yu, J., Tan, E., Wang, L., Huang,
  Q., Wang, Y., Mahoney, M., et~al.
\newblock Hawq-v3: Dyadic neural network quantization.
\newblock In \emph{International Conference on Machine Learning}, pp.\
  11875--11886. PMLR, 2021.

\bibitem[Zachariadis et~al.(2020)Zachariadis, Satpute, G{\'o}mez-Luna, and
  Olivares]{spgemm2020}
Zachariadis, O., Satpute, N., G{\'o}mez-Luna, J., and Olivares, J.
\newblock Accelerating sparse matrix--matrix multiplication with gpu tensor
  cores.
\newblock \emph{Computers \& Electrical Engineering}, 88:\penalty0 106848,
  2020.

\bibitem[Zheng et~al.(2020)Zheng, Jia, Sun, Wu, Yu, Haj-Ali, Wang, Yang, Zhuo,
  Sen, et~al.]{ansor2020}
Zheng, L., Jia, C., Sun, M., Wu, Z., Yu, C.~H., Haj-Ali, A., Wang, Y., Yang,
  J., Zhuo, D., Sen, K., et~al.
\newblock Ansor: Generating high-performance tensor programs for deep learning.
\newblock In \emph{14th $\{$USENIX$\}$ Symposium on Operating Systems Design
  and Implementation ($\{$OSDI$\}$ 20)}, pp.\  863--879, 2020.

\bibitem[Zhu et~al.(2019)Zhu, Zhang, Gu, and Xie]{sparsetc2019}
Zhu, M., Zhang, T., Gu, Z., and Xie, Y.
\newblock Sparse tensor core: Algorithm and hardware co-design for vector-wise
  sparse neural networks on modern gpus.
\newblock In \emph{Proceedings of the 52nd Annual IEEE/ACM International
  Symposium on Microarchitecture}, pp.\  359--371, 2019.

\end{thebibliography}
\bibliographystyle{icml2021}

\end{document}